\documentclass[12pt,a4paper]{cibb}

\usepackage{subcaption,graphicx}
\usepackage{amsmath,amsfonts,latexsym,amssymb,euscript,xr}
\usepackage{adjustbox}
\usepackage{booktabs}
\usepackage[nodayofweek]{datetime}
\usepackage{hyperref}
\usepackage{multirow}
\usepackage{soul}

\usepackage[table]{xcolor}
\usepackage{color,colortbl,tabularx}

\usepackage[english]{babel}
\usepackage[protrusion=true,expansion=true]{microtype}
\usepackage{amsmath,amsfonts,amsthm}
\usepackage{pifont}
\usepackage{comment}

\usepackage{caption}
\usepackage[font={small}]{caption}

\definecolor{LightBlue}{rgb}{0.88,0.9,0.9}

\title{\Large $\ $\\ \bf Exploring the Impact of Environmental Pollutants on Multiple Sclerosis Progression}

\author{\large Elena Marinello$^1$, Erica Tavazzi$^{1}$, Enrico Longato$^{1}$, Pietro Bosoni$^{2}$, Arianna Dagliati$^{2}$, Mahin Vazifehdan$^{2}$, Riccardo Bellazzi$^{2}$, Isotta Trescato$^1$, Alessandro Guazzo$^1$, Martina Vettoretti$^1$, Eleonora Tavazzi$^{3}$, Lara Ahmad$^{3}$, Roberto Bergamaschi$^{3}$, Paola Cavalla$^{4}$, Umberto Manera$^{4}$, Adriano Chiò$^{4}$, Barbara Di Camillo$^{1,5,*}$}
\address{\footnotesize $\ $\\$^1$ Department of Information Engineering, University of Padova, Padova, Italy. \\
$^2$  Department of Industrial and Information Engineering, University of Pavia, Pavia, Italy. \\ 
$^3$ Multiple Sclerosis Centre, IRCCS Mondino Foundation, Pavia, Italy. \\
$^4$ Department of Neurosciences "Rita Levi Montalcini", University of Turin, Torino, Italy. \\
$^5$ Department of Comparative Biomedicine and Food Science, University of Padova, Padova, Italy. \\
$^*$corresponding author (barbara.dicamillo@unipd.it)
}

\abstract{\small Multiple Sclerosis, Relapse Prediction, Environmental Data, Pollutants, Machine Learning. \normalsize
\\[17pt]
{\bf Abstract.}  
Multiple Sclerosis (MS) is a chronic autoimmune and inflammatory neurological disorder characterised by episodes of symptom exacerbation, known as relapses. In this study, we investigate the role of environmental factors in relapse occurrence among MS patients, using data from the H2020 BRAINTEASER project. We employed predictive models, including Random Forest (RF) and Logistic Regression (LR), with varying sets of input features to predict the occurrence of relapses based on clinical and pollutant data collected over a week. The RF yielded the best result, with an AUC-ROC score of 0.713. Environmental variables, such as precipitation, NO\textsubscript{2}, PM\textsubscript{2.5}, humidity, and temperature, were found to be relevant to the prediction.}

\begin{document}
\thispagestyle{myheadings}
\pagestyle{myheadings}
\markright{\tt Proceedings of CIBB 2024}

\vspace{-5pt}
\section{\bf Introduction}
Multiple Sclerosis (MS) is a chronic autoimmune, inflammatory neurological disease affecting the central nervous system, with a highly variable progression, which often manifests with episodes of reversible neurological impairments, followed by a gradual decline in neurological function over time ~\cite{pmid22605909}. The periods when symptoms get worse are known as relapses (or exacerbations), while periods when symptoms improve or disappear are known as remissions. Various risk factors have been linked to the prognosis of MS, including stress ~\cite{pmid21335982}, low vitamin D levels ~\cite{pmid28320120}, and environmental factors, especially particulate matter (PM) concentrations ~\cite{pmid33316725}. 

In this study, we focused on the influence of pollutants and environmental factors on relapses connected with MS progression. For this purpose, we employed a dataset of clinical information of MS patients enriched with environmental data collected through air-quality sensors, collected in the context of the H2020 BRAINTEASER (``Bringing Artificial Intelligence home for a better care of amyotrophic lateral sclerosis and multiple sclerosis'') project. Specifically, we developed a portfolio of models to predict the occurrence vs. non-occurrence of a relapse, based on the clinical and environmental information collected over a week, considering one week of data as possibly predictive of the occurrence of a relapse in the following week. This is consistent with multiple studies (e.g., ~\cite{pmid28407574}) where the exposition to pollutants, lagging from 1 to 3 days, was found to be associated with an increased risk of relapses.
We tested different configurations of the input features with a linear method (Logistic Regression, LR) and a non-linear method, (Random Forest, RF). 
The best performing model, with an AUC-ROC score of 0.713 and an AUC-PR of 0.639, was an RF model using a set of automatically-selected features, including environmental variables such as precipitation, NO\textsubscript{2}, PM\textsubscript{2.5}, humidity, and temperature, along with key clinical variables such as time since onset, age at onset, and diagnostic delay. 

\vspace{-5pt}
\section{\bf Materials}

The data used in this study were part of the dataset from the H2020 BRAINTEASER project and collected in two MS clinical centers located in Italy: the Fondazione Istituto Neurologico Nazionale C. Mondino of Pavia and the Città della Salute e della Scienza of Torino \cite{faggiolibrainteaser}. The data include clinical information, relative to demographic data (such as sex, ethnicity, residence), clinical information at onset (such as diagnostic delay and symptoms), dynamic assessments of disability status (Expanded Disability Status Scale, EDSS score, and subscores related to different functional systems), and pharmacological treatments (such as Anatomical Therapeutic Chemical, ATC codes, use as first/second-line treatment) for MS patients, spanning from January 1st, 2013 to January 11th, 2021. An innovative aspect was the augmentation of patient information with environmental factors. To decide which environmental factor to include, the WHO's new guidelines on air quality levels were considered, given the evidence on health effects from exposure to PM pollutants and chemical agents \cite{whoGlobalQuality}. Therefore, the data include PM pollutants (PM\textsubscript{2.5} and PM\textsubscript{10}), chemical agents (CO, NO\textsubscript{2}, etc.), and weather conditions (such as humidity and temperature). Air pollutant data from public monitoring stations were collected daily from the European Air Quality Portal \cite{europaEuropeanQuality}. Similarly, weather data were gathered daily from the European Climate Assessment \& Dataset station network \cite{Climate}. To compute patients' exposure, the geographical coordinates (longitude and latitude) of each measurement station were matched to specific postcodes, identifying the nearest station to each patient's residence postcode. Subsequently, daily measurements for each air pollutant and weather factor were aggregated as averages into consecutive weeks starting from January 1, 2013.
This vast collection of information allowed us to develop predictive models that consider clinical and environmental data together. 

To limit the number of variables to be included in the model while preserving their information content, we performed a correlation analysis between the available environmental variables.
A high correlation was observed between the two variables of PM and CO, NO\textsubscript{2}, SO\textsubscript{2}, humidity, and sea-level pressure. Additionally, we detected a high correlation within the group of variables consisting of O\textsubscript{3}, global radiation, and the minimum/maximum/average temperatures. Based on these considerations, we selected PM\textsubscript{10}, NO\textsubscript{2}, wind speed, humidity, precipitation, and minimum/maximum/average temperature as predictors. 

We then structured our data for predicting the occurrence (vs. non-occurrence) of a relapse in a given week starting from the variables collected over the previous week. 
For each subject with at least one relapse recorded over the observation interval, we extracted the first relapse for which a one-week-long window of previous data was available. These subjects, associated with the outcome ``occurrence of relapse = yes'', were named as the \textit{cases} for our study. 
We then extracted the cohort of subjects who did not experience any relapse in our observation interval. From this group, we selected, for each case, a matched counterpart who did not experience a relapse in the same week of the year (regardless of calendar year) and who had a week of clinical and environmental data recorded for the previous week. This design choice allowed us to correct for possible seasonal effects.
We also observed that, in the observation period covered by the retrospective BRAINTEASER data, new and more effective Disease Modifying Therapies (DMTs) became available -- in particular, starting from 2018, with the introduction of cladribine ~\cite{pmid30050387}. Therefore, we added an additional constraint to the definition of the counterparts, matching them not only on the week of the year of the corresponding case, but also on the availability of these advanced DMTs (practically, not cross-matching the subjects before and after 2018). In this way, we mitigated the potential decrease in the baseline relapse risk after 2018, which may have been due to the efficacy of these drugs. The subjects selected with this matching procedure, associated with the outcome ``occurrence of relapse = no'', were named as the \textit{controls} for our study.
The resulting dataset comprised 409 cases and 393 controls, with a total of 802 subjects. 16 relapse patients did not have a matched control, due to the impossibility of matching.

From this dataset, we removed all the variables exceeding 30\% of missing values, the information on pharmacological treatments, and computed the derived variables \textit{time since onset} (corresponding to the difference in days between the onset and the week used for prediction) and \textit{PM\textsubscript{10} ratio} and \textit{NO\textsubscript{2} ratio}, computed as the estimate of the fraction of days in the given week in which the mean value of the pollutant was above the WHO safety thresholds. Four of these variables, i.e., \textit{ethnicity}, \textit{residence classification}, and \textit{season}, were eventually made dummy deleting their most frequent level, finally obtaining a dataset consisting of 36 predictor variables. 
The first column of Table \ref{tab:TabFeatures} reports a summary of the variables considered after the preprocessing. 
We then split the dataset into a training (70\%) and a test (30\%) sets stratified by the outcome, resulting in, respectively, 561 and 241 subjects.

    \begin{table}[h!]
    \centering
    \resizebox{\textwidth}{!} {
    \begin{tabular}{|l|l|c|c|c|c|}
    \hline
    \rowcolor[HTML]{EFEFEF} 
    \cellcolor[HTML]{EFEFEF}{\color[HTML]{000000} }   & \cellcolor[HTML]{EFEFEF}{\color[HTML]{000000} }                                     & \cellcolor[HTML]{EFEFEF}{\color[HTML]{000000} }                                    & \multicolumn{2}{c|}{\cellcolor[HTML]{EFEFEF}{\color[HTML]{000000} \textbf{\begin{tabular}[c]{@{}c@{}}With features \\ selected by RF\end{tabular}}}} & \cellcolor[HTML]{EFEFEF}{\color[HTML]{000000} }                                   \\ \cline{4-5}
    \rowcolor[HTML]{EFEFEF} 
    \multirow{-3}{*}{\cellcolor[HTML]{EFEFEF}{\color[HTML]{000000} \textbf{\begin{tabular}[c]{@{}c@{}} Category \end{tabular}}}} & \multirow{-3}{*}{\cellcolor[HTML]{EFEFEF}{\color[HTML]{000000} \textbf{\begin{tabular}[c]{@{}c@{}} Variable name\end{tabular}}}} & \multirow{-2}{*}{\cellcolor[HTML]{EFEFEF}{\color[HTML]{000000} \textbf{\begin{tabular}[c]{@{}c@{}}Removing \\ correlated \\ features \newline \end{tabular}}}} & \multicolumn{1}{c|}{\cellcolor[HTML]{EFEFEF}\textbf{Selected}}  & \textbf{\begin{tabular}[c]{@{}c@{}}Feature \\ Importance\end{tabular}}  &  \textbf{\begin{tabular}[c]{@{}c@{}}Backward \\ Feature\\ Selection\newline \end{tabular}} \\
    \hline
    \hline  Clinical (current week) & Time since onset & $x$ & $x$ & 0.113750 & $x$ \\
    \hline  Demographic & Age at onset & $x$ & $x$ & 0.082736 & $x$ \\
    \hline  Environmental & Wind speed mean & $x$ & $x$ & 0.082650 & \\
    \hline  Environmental & Precipitation mean & & $x$ & 0.068039 & $x$ \\
    \hline  Environmental & NO\textsubscript{2} mean & $x$ & $x$ & 0.067042 & \\
    \hline  Environmental & Average temperature mean & $x$ & $x$ & 0.065835 & \\
    \hline  Environmental & Max temperature mean & & $x$ & 0.065355 & $x$ \\
    \hline  Clinical (at onset) & Diagnostic delay & $x$ & $x$ & 0.064868 & $x$ \\
    \hline  Environmental & Humidity mean & & $x$ & 0.059956 & $x$ \\
    \hline  Environmental & Min temperature mean & & $x$ & 0.058756 & $x$ \\
    \hline  Environmental & PM\textsubscript{10} mean & $x$ & $x$ & 0.054591 & \\
    \hline  Clinical (most recent visit) & EDSS score & $x$ & $x$ & 0.035765 & $x$ \\
    \hline  Demographic & Sex & $x$ & & & $x$ \\
    \hline  Clinical (at onset) & MS in pediatric age & $x$ & & & \\
    \hline  Clinical (at onset) & Spinal cord symptoms at onset & $x$ & & & \\
    \hline  Clinical (at onset) & Brainstem symptoms at onset & $x$ & & & $x$ \\
    \hline  Clinical (at onset) & Eye symptoms at onset & $x$ & & & \\
    \hline  Clinical (at onset) & Supratentorial symptoms at onset & $x$ & & & \\
    \hline  Clinical (at onset) & Other symptoms at onset & $x$ & & & $x$ \\
    \hline  Clinical (most recent visit) & Pyramidal functional system score & & & &\\
    \hline  Clinical (most recent visit) & Cerebellar functional system score & & & &\\
    \hline  Clinical (most recent visit) & Brainstem functional system score & & & & \\
    \hline  Clinical (most recent visit) & Sensory functional system score & & & & $x$ \\
    \hline  Clinical (most recent visit) & Bowel and bladder functional system score & & & & $x$ \\
    \hline  Clinical (most recent visit) & Visual function score & & & &\\
    \hline  Clinical (most recent visit) & Cerebral functions score & & & & \\
    \hline  Clinical (most recent visit) & Ambulation functional system score & & & & $x$ \\
    \hline  Environmental & PM\textsubscript{10} ratio & & & &\\
    \hline  Environmental & NO\textsubscript{2} ratio & & & & \\
    \hline  Demographic & Caucasian ethnicity* & & & & \\
    \hline  Demographic & Black-african ethnicity & $x$ & & & \\
    \hline  Demographic & Hispanic ethnicity & $x$ & & & \\
    \hline  Demographic & Residence in towns* & & & & \\
    \hline  Demographic & Residence in cities & $x$ & & & $x$ \\
    \hline  Demographic & Residence in rural area & $x$ & & & \\
    \hline  Environmental & Summer season* & & & & \\
    \hline  Environmental & Spring season & & & & \\
    \hline  Environmental & Autumn season & & & & \\
    \hline  Environmental & Winter season & & & & $x$\\
    \hline
    \end{tabular}
    }
    \caption{\textbf{Table of features selected by the different methods.} Rows marked with * correspond to the most frequent level of the corresponding categorical variable, which were removed after the conversion into dummy variables. 
    The presence of an $x$ indicates that the feature was selected by the method.   
    \label{tab:TabFeatures}}
    \vspace{-4mm}
    \end{table} 

\vspace{-5pt}
\section{\bf Methods}
Since the relationships between the variables were not known a priori, we explored models of different types and capacities.
Specifically, we implemented one linear method, the Logistic Regression (LR), and one non-linear method, the Random Forest (RF) model. We employed a 4-fold cross-validation (CV) procedure to estimate the hyperparameters of the models. Within each CV loop and for the training of the final model, the imputation of missing values in the categorical variables was done by replacing missing values with the most frequent value in each column, while the numerical variables were imputed via multivariate imputation by chained equations (MICE) method  
using the 3 most correlated features. Then, the numerical features were standardized. Moreover, different feature selection mechanisms were incorporated into specific pipelines, later described in detail.

Each model was trained using the specified hyperparameters within each iteration of the CV loop, and the optimal hyperparameters were determined via grid search. The hyperparameter for the LR was the inverse of the regularisation strength \textit{C} (range: [0.01, 0.1, 1, 10, 100]). As for the RF, the hyperparameters were whether to use bootstrap resampling when building trees (\textit{bootstrap} True or False), the maximum number of features at each split \textit{max\_features} (range: ['sqrt']), the minimum number of elements in a leaf \textit{min\_samples\_leaf} (range: [2, 4, 8, 18]), and the number of weak estimators \textit{n\_estimators} (range: [50, 100, 200, 350, 500, 650, 800, 950]).  Finally, the model performance was evaluated using the Area Under the Receiver Operating Characteristic Curve (AUC-ROC) and the Area Under the Precision-Recall Curve (AUC-PR). Moreover, the confidence intervals (CI) were calculated to give a measure of the uncertainty associated with the AUC-ROC score.

We tested 3 different feature selection approaches, which we compared with the full models. The first approach consisted of the manual removal of the correlated features in the training set (correlation coefficient $>0.3$). The second feature selection method involved using the RF's Variable Importance in Projection (VIP) metric to rank and order variables based on their feature importance scores, derived from the reduction in impurity; features with importance scores higher than the mean were selected for both RF and LR model training. The final setting was the implementation of a Backward Feature Selection (BFS) technique based on the highest AUC-ROC score. This was only applied to the LR for computational reasons.

The analysis was conducted using the scikit-learn library (v.1.4.2) in Python (v.3.12.2).

\vspace{-10pt}
\section{\bf Results and Discussion}
~\autoref{tab:TabPerformance} shows the performance of the different models on the test set, while ~\autoref{tab:TabFeatures} reports the features selected with the three different feature selection methods. 

\begin{table}[h!] 
\centering
\resizebox{\textwidth}{!} {
\begin{tabular}{c|l|c|c|c|c}
\toprule[1pt]
\multicolumn{2}{l}{} & \textbf{Using all features}                         & \textbf{Removing correlated features} & \textbf{With features selected by RF} & \textbf{Backward Feature Selection}                  
\\ \hline
\rowcolor{LightBlue} 
\multicolumn{1}{c|}{\cellcolor{LightBlue}}           & \multicolumn{1}{l|}{\cellcolor{LightBlue}\textbf{AUC-ROC score}} & \begin{tabular}[c]{@{}c@{}}0.708 \\ {[}0.637, 0.771{]}\end{tabular} & \begin{tabular}[c]{@{}c@{}}0.701 \\ {[}0.629, 0.763{]}\end{tabular} & \textbf{\begin{tabular}[c]{@{}c@{}}0.713 \\ {[}0.642, 0.776{]}\end{tabular}} & - \\ \cline{2-6} 
\rowcolor{LightBlue} 
\multicolumn{1}{c|}{\multirow{-2}{*}{\cellcolor{LightBlue}\textbf{Random Forest}}}& \multicolumn{1}{l|}{\cellcolor{LightBlue}\textbf{AUC-PR score}} & 0.666 & 0.694& 0.639& -
\\ \hline
\rowcolor[HTML]{FFFFFF} 
\multicolumn{1}{c|}{\cellcolor[HTML]{FFFFFF}}           & \multicolumn{1}{l|}{\cellcolor[HTML]{FFFFFF}\textbf{AUC-ROC score}} & \begin{tabular}[c]{@{}c@{}}0.686 \\ {[}0.615, 0.749{]}\end{tabular} & \begin{tabular}[c]{@{}c@{}}0.682 \\ {[}0.613, 0.746{]}\end{tabular} & \begin{tabular}[c]{@{}c@{}}0.696 \\ {[}0.626, 0.762{]}\end{tabular}      & \begin{tabular}[c]{@{}c@{}}0.697 \\ {[}0.628, 0.763{]}\end{tabular} \\ \cline{2-6} 
\rowcolor[HTML]{FFFFFF} 
\multicolumn{1}{c|}{\multirow{-2}{*}{\cellcolor[HTML]{FFFFFF}\textbf{Logistic Regression}}} & \multicolumn{1}{l|}{\cellcolor[HTML]{FFFFFF}\textbf{AUC-PR score}} & 0.646                & 0.604 & 0.628 & 0.639 \\ \hline
\end{tabular}
}
\caption{\textbf{Model performance on the test set}.
    Area Under the Receiver Curve (AUC-ROC) scores and its Confidence Intervals (CI), and Area Under the Precision-Recall Curve (AUC-PR). The reference value for the AUC-PR is 0.527. The best performance in terms of AUC-ROC is highlighted in bold. \label{tab:TabPerformance}}
\vspace{-4mm}
\end{table}

\begin{figure}[htbp]
    \vspace{-10pt}
     \centering
     \includegraphics[width=0.5\textwidth]{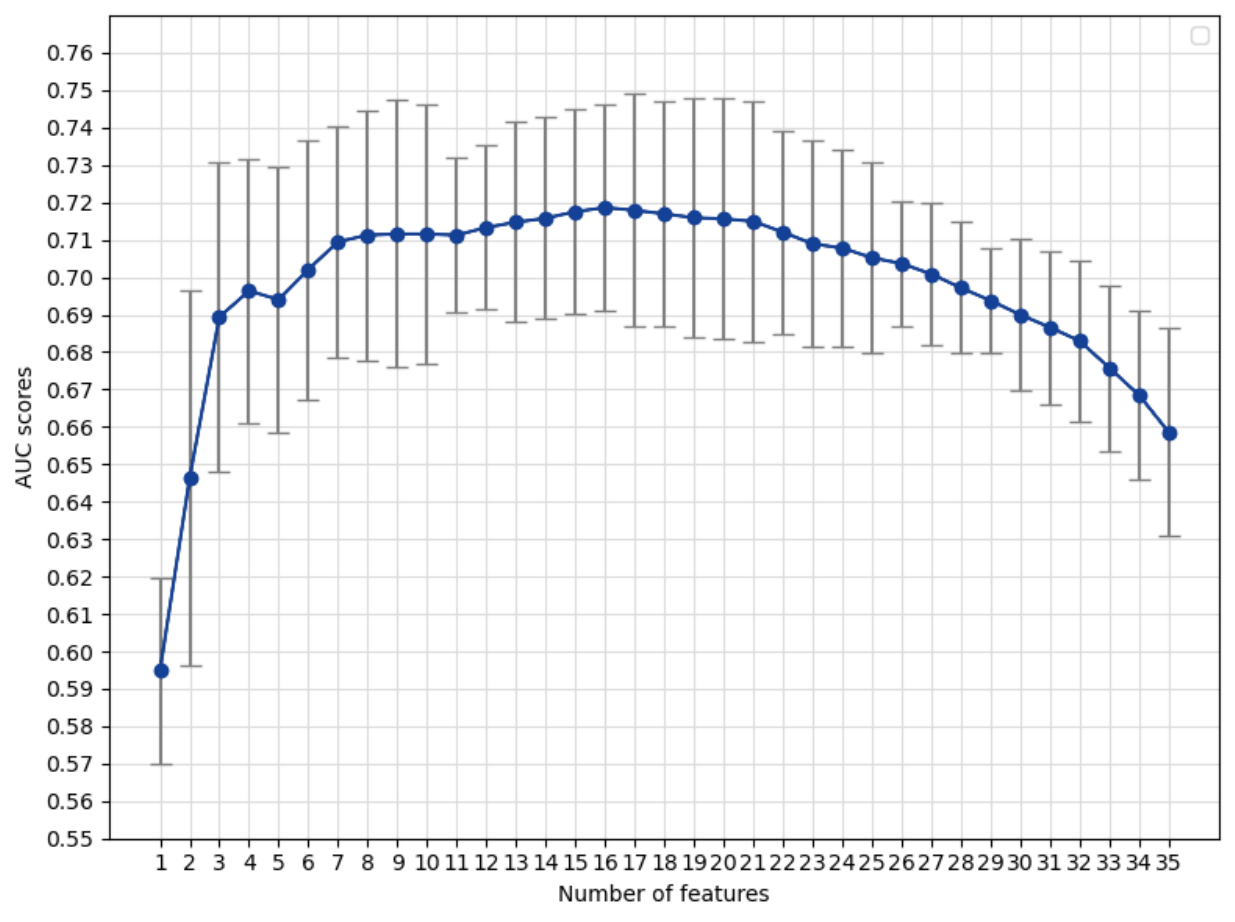}
    \caption{\textbf{AUC-ROC as a function of the number of features in Backward Feature Selection}, presented as mean and standard deviation.
    \label{fig:Fig13}}
    \vspace{-20pt}
    \end{figure}

In the analysis using all features, the RF exhibited optimal performance with the following hyperparameters: \textit{n\_estimators}: 100, \textit{min\_samples\_leaf}: 4, \textit{max\_features}: sqrt, \textit{bootstrap}: True. The AUC-ROC for the RF model was 0.708, (95\% CI: 0.637 - 0.771), while the LR model yielded an AUC-ROC of 0.686, (95\% CI: 0.615 - 0.749) with the best hyperparameter \textit{C} equal to 10. With respect to the setting where the correlated features were removed, we chose the variables reported in the third column of ~\autoref{tab:TabFeatures}, which included, among others: \textit{PM\textsubscript{10} mean}, representing the most informative variable (in terms of non-missing information) among the two PM agents; \textit{NO\textsubscript{2} mean}, representing the most informative chemical agent in terms of non-missing values; \textit{wind speed mean}, representing atmospheric agents and being the only uncorrelated atmospheric agent; and \textit{EDSS score} which was chosen due to its clinical relevance.
After applying this feature selection scheme based on the correlation matrix, the hyperparameters selected for the RF resulted in an optimal configuration with \textit{n\_estimators}:100, \textit{min\_samples\_leaf}:4, \textit{max\_features}: sqrt, and \textit{bootstrap}: True. The AUC-ROC for the RF model was 0.701 (95\% CI: 0.629 - 0.763). On the other hand, the AUC-ROC score for the LR model was 0.682 (95\% CI: 0.613 - 0.746). The best \textit{C} value was 0.1. In the third analysis, we used an RF model to select the most important features according to the VIP metric. The ranking of all features and their importance is reported in ~\autoref{tab:TabFeatures}, fourth and fifth columns, respectively. The features selected for the final model were those with feature importance scores higher than the mean importance (0.027). 
The AUC-ROC of the RF model trained using the feature selected by RF only was equal to 0.713 (95\% CI: 0.642 - 0.776). The optimal hyperparameters were \textit{n\_estimators}: 500, \textit{min\_samples\_leaf}: 2, \textit{max\_features}: 'sqrt', and \textit{bootstrap}: True.
On the other hand, the LR using the features selected by the RF yielded the following results: the best \textit{C} value was 10 and the AUC-ROC score resulted in 0.696 (95\% CI: 0.626 - 0.762). With respect to the BFS method, ~\autoref{fig:Fig13} reports the plot of cross-validation AUC-ROC vs. the number of features. The best subset of features, corresponding to the maximum of the curve, comprised 16 variables, which are reported in the last column of ~\autoref{tab:TabFeatures}.
The AUC-ROC of the corresponding LR model, reported in 
\autoref{tab:TabPerformance}, was 0.697 (95\% CI: 0.628 – 0.763), with the best hyperparameter \textit{C} equal to 10. 	

\vspace{-7pt}
\section{Conclusion}
In this work, we approached the problem of predicting the occurrence vs. non-occurrence of a relapse in MS starting from the clinical and environmental variables collected over a week. We tested different configurations of the input features, selected via different manual and automatic methods, developing different versions of LR and RF models.
We obtained the best performance, as measured by the AUC-ROC, with the RF feature selection method and a subsequent reapplication of the RF model resulting in an AUC-ROC score of 0.713 and a satisfactory AUC-PR of 0.639.
The selected features included key clinical variables, i.e.,  \textit{time since onset}, \textit{age at onset}, \textit{diagnostic delay}, and \textit{EDSS score}. In addition, the environmental variables  \textit{wind speed mean}, \textit{precipitation mean}, \textit{NO\textsubscript{2} mean}, \textit{average temperature mean}, \textit{max temperature mean}, \textit{humidity mean}, \textit{min temperature mean}, and \textit{PM\textsubscript{10} mean} were also selected, suggesting their importance for the prediction of the relapse. 
Interestingly, it appears that the best set of features selected via a data-driven method such as BFS was a combination of clinical information that was a proxy of disease progression relative to the onset, and of meaningful variables, collected immediately prior to the relapse, which were identified by the WHO as negatively impacting an idividual's health state. Considering these findings, we can conclude that MS relapses can be predicted with acceptable performance with one week of environmental and clinical data, with the former type of variables being selected via multiple manual and automatic feature selection approaches. 

\vspace{-5pt}
\section*{\bf Conflict of interests}
\label{sec:CONFLICT-OF-INTERESTS}
None declared.

\section*{\bf Funding}
\label{sec:FUNDING}
This research has been partially supported by the Horizon 2020 project BRAINTEASER (Bringing Artificial Intelligence home for a better care of amyotrophic lateral sclerosis and multiple sclerosis) funded from the European Union’s Horizon 2020 research and innovation programme under grant agreement No. GA101017598..

\section*{\bf Availability of data}
\label{sec:AVAILABILITY}
The retrospective BRAINTEASER MS data used in this work are available on Zenodo: \url{https://doi.org/10.5281/zenodo.8083181}.

\footnotesize
\bibliographystyle{unsrt}
\bibliography{main.bib} 

\begin{thebibliography}{10}

\bibitem{pmid22605909}
M.~M. Goldenberg.
\newblock {{M}ultiple sclerosis review}.
\newblock {\em P T}, 37(3):175--184, Mar 2012.

\bibitem{pmid21335982}
A.~K. Artemiadis and et~al.
\newblock {{S}tress as a risk factor for multiple sclerosis onset or relapse: a systematic review}.
\newblock {\em Neuroepidemiology}, 36(2):109--120, 2011.

\bibitem{pmid28320120}
C.~Hartl and et~al.
\newblock {{S}easonal variations of 25-{O}{H} vitamin {D} serum levels are associated with clinical disease activity in multiple sclerosis patients}.
\newblock {\em J Neurol Sci}, 375:160--164, Apr 2017.

\bibitem{pmid33316725}
A.~Scartezzini and et~al.
\newblock {{A}ssociation of {M}ultiple {S}clerosis with {P}{M} 2.5 levels. {F}urther evidence from the highly polluted area of {P}adua {P}rovince, {I}taly}.
\newblock {\em Mult Scler Relat Disord}, 48:102677, Feb 2021.

\bibitem{pmid28407574}
J.~Roux and et~al.
\newblock {Air pollution by particulate matter PM10 may trigger multiple sclerosis relapses.}
\newblock {\em Environ Res}, 156:404--410, Jul 2017.

\bibitem{faggiolibrainteaser}
G.~Faggioli et~al.
\newblock {BRAINTEASER ALS and MS Datasets. Zenodo, June 2023}.

\bibitem{whoGlobalQuality}
{W}{H}{O} global air quality guidelines: particulate matter ({P}{M}2.5 and {P}{M}10);, ozone, nitrogen dioxide, sulfur dioxide and carbon monoxide --- iris.who.int.
\newblock \url{https://iris.who.int/handle/10665/345329}.
\newblock [Accessed 07-07-2024].

\bibitem{europaEuropeanQuality}
{E}uropean {A}ir {Q}uality {P}ortal; {T}he {D}ata and {I}nformation {G}ateway on {A}ir {Q}uality in {E}urope --- aqportal.discomap.eea.europa.eu.
\newblock \url{https://aqportal.discomap.eea.europa.eu/}.
\newblock [Accessed on 22-05-2024].

\bibitem{Climate}
{C}opernicus {C}limate {D}ata {S}tore | {C}opernicus {C}limate {D}ata {S}tore --- cds.climate.copernicus.eu.
\newblock \url{https://cds.climate.copernicus.eu/cdsapp#!/dataset/insitu-gridded-observations-europe?tab=overview}.
\newblock [Accessed on 22-05-2024].

\bibitem{pmid30050387}
A.~N. Boyko and O.~V. Boyko.
\newblock {{C}ladribine tablets' potential role as a key example of selective immune reconstitution therapy in multiple sclerosis}.
\newblock {\em Degener Neurol Neuromuscul Dis}, 8:35--44, 2018.

\end{thebibliography}
\normalsize

\end{document}